\documentclass{article}

     \usepackage[final]{neurips_2020}

\usepackage[utf8]{inputenc} % allow utf-8 input
\usepackage[T1]{fontenc}    % use 8-bit T1 fonts
\usepackage{hyperref}       % hyperlinks
\usepackage{url}            % simple URL typesetting
\usepackage{booktabs}       % professional-quality tables
\usepackage{amsfonts}       % blackboard math symbols
\usepackage{nicefrac}       % compact symbols for 1/2, etc.
\usepackage{microtype}      % microtypography
\usepackage{xcolor}         % colors
\usepackage{amsmath}
\usepackage{algorithm}
\usepackage{multicol}
\usepackage{multirow}
\usepackage{algpseudocode}
\usepackage{dsfont}

\usepackage{amsmath,amsthm,amssymb,amsfonts}
\usepackage{graphicx}
\usepackage{xcolor}
\usepackage{hyperref}
\usepackage{soul}
\usepackage{color}

\newtheorem{definition}{Definition}

\title{An End-to-End Network Pruning Pipeline with Sparsity Enforcement}

\author{%
  Evan Dogariu \\
  Department of Computer Science\\
  Princeton University\\
  \texttt{edogariu@princeton.edu} \\
}

\begin{document}

\maketitle

\begin{abstract}
    Neural networks have emerged as a powerful tool for solving complex tasks across various domains, but their increasing size and computational requirements have posed significant challenges in deploying them on resource-constrained devices. Neural network sparsification, and in particular pruning, has emerged as an effective technique to alleviate these challenges by reducing model size, computational complexity, and memory footprint while maintaining competitive performance. However, many pruning pipelines modify the standard training pipeline at only a single stage, if at all. In this work, we look to develop an end-to-end training pipeline that befits neural network pruning and sparsification at \textbf{all} stages of training. To do so, we make use of nonstandard model parameter initialization, pre-pruning training methodologies, and post-pruning training optimizations. We conduct experiments utilizing combinations of these methods, in addition to different techniques used in the pruning step, and find that our combined pipeline can achieve significant gains over current state of the art approaches to neural network sparsification.
\end{abstract}

\section{Introduction}
Neural networks have revolutionized the field of machine learning by achieving state-of-the-art performance in various complex tasks. However, the growing demand for deploying these models on resource-constrained devices, such as mobile phones, embedded systems, and Internet of Things (IoT) devices, makes it valuable to investigate methods to decrease inference time and memory footprint without sacrificing performance. Various techniques that fall under the category of neural network sparsification have been developed to tackle this problem, and one of the most promising subfields is that of neural network pruning. 
\newline \newline
The goal of pruning is to determine which subset of the network parameters are needed for good performance, and to retrain this subnetwork. This standard pruning pipeline is motivated by the empirical support for the Lottery Ticket Hypothesis \cite{lth}, which suggests that within an overparameterized neural network, there exist sparse subnetworks, or "winning tickets," that can achieve comparable performance to the original dense network when trained in isolation. This groundbreaking work demonstrated the potential for pruning to discover highly efficient neural network architectures without losing accuracy.
\section{Related Works}
In this section, we relate various prior methods for network pruning and  for the training of sparse neural networks. We will make use of these techniques as building blocks of our pipeline, baselines to compare against, and motivations for our research.

\subsection{Neural Network Pruning}
Neural network pruning is a technique that involves selectively removing connections, neurons, or filters from a trained neural network, resulting in a sparser architecture. More precisely, given a neural network $f(x, \theta)$ parameterized by $\theta$, the objective is to generate a mask $M \in \{0, 1\}^{|\theta|}$ (here, $|\theta|$ is the number of parameters) with the following two desirable properties:
\begin{enumerate}
    \item the mask is \textit{sparse}; that is, $||M||_0 \ll |\theta|$ (here, $||\cdot||_0$ is the $L^0$ "norm", measuring the number of nonzero elements)
    \item applying the mask to the parameters of the network \textit{doesn't compromise performance}; that is, the model $f(x, \theta \odot M)$ performs well (here, $\odot$ is the element-wise product)
\end{enumerate}

% \subsubsection{Model-Based Pruning Techniques}
Techniques for building the mask $M$ that make use of training a model are called \textbf{training-based}. In such methods, we first train a model $f(x, \theta)$ to convergence, and then apply our pruning method to this trained model (perhaps iteratively). These methods are generally the most common, and also what motivated the Lottery Ticket Hypothesis \cite{lth}. There are also newer methods for constructing the mask $M$ that do not require any learning updates on the model, but are instead based solely on the model's architecture and the initialization of the parameters $\theta$; we call these methods \textbf{training-free}. We recount methods of both types below.
\newline \newline
\textbf{Magnitude-Based Pruning.} The main method in this class is \textit{magnitude-based pruning}, in which weights with the lowest magnitude are assumed to be negligible and can be pruned away \cite{liang2021pruning}; importantly, this method is one-shot and only requires one round of training. Furthermore, the technique is interpretable, efficient, and easy to implement.
\newline \newline
\textbf{Iterative Pruning.} Another technique is \textit{iterative pruning}, in which magnitude-based pruning is applied in rounds, pruning a small subset of the weights each iteration \cite{paganini2020iterative}. This approach can often demonstrate better performance than a single round of magnitude-based pruning, but is much more computationally burdensome. 
 \newline \newline \textbf{Random Pruning.} In \textit{random pruning}, networks are pruned randomly at initialization. This is the most naive way to approach network pruning, and generally has poor performance.
\newline \newline \textbf{SNIP.} In \textit{single-shot network pruning} (SNIP) \cite{snip}, networks are pruned at initialization based on a saliency criterion that determines structurally important connections. The motivation is that connections that are architecturally important on initialization become the connections that are needed in the trained model.
\newline \newline
\textbf{GraSP.} In \textit{gradient signal preservation} (GraSP) \cite{grasp}, networks are pruned at initialization in a fashion that preserves the overall gradient flow through the network. The idea here is that the gradient flow is what dominates training trajectories, and so weights that don't contribute to the gradient flow will not be learned well.
\newline \newline \textbf{SynFlow.} In \textit{iterative synaptic flow pruning} (SynFlow) \cite{synflow}, weights are pruned on initialization in a way that preserves the total flow of synaptic strengths, avoiding the layer collapse that comes about from gradient-based methods.
\newline \newline

\subsection{Sparse Training Methods}
Below, we mention several works that are tailored specifically to training neural networks that either are or will be sparse.
% \subsubsection{Pre-Pruning Training Techniques}
% Because of the desirability of successful network sparsification and pruning, there has been much research into methods with which to train neural networks that promote or enforce sparsity, which can be applied in the pre-pruning stage. 
\newline \newline
\textbf{ZerO Initialization.} One such method is the \textit{ZerO initialization} \cite{zhao2022zero}, which makes use of a theoretically-justified deterministic initialization that begins as low rank as possible. This sharply deviates from the tried and true practice of random initialization, which begins with maximal rank transformations at each layer, but it has been empirically shown to produce networks that can be more effectively pruned. 
\newline \newline
\textbf{Learnable Masks.} Another approach, detailed in \cite{srinivas2016training}, is to learn a sparsity mask alongside the weights of the model, and to prune according to this learned mask. During the pre-pruning training, an extra regularization term is added to the loss to ensure that the learned mask is indeed sparse. This technique, while certainly improving pruning capabilities, effectively doubles the memory footprint of the pre-pruning training. However, later stages of training and the resulting final model do not suffer this memory penalty.
% \subsubsection{Post-Pruning Training Techniques}
% The post-pruning training stage requires the training of an already-sparse model, for which the training dynamics and design decisions look different from the setting. As such, several techniques have been developed that are specialized to this mode of training.
\newline \newline
\textbf{ToST.} The sparse training toolkit \textit{ToST} \cite{jaiswal2022training} is a collection of small optimizations and changes to the traditional training methodology that is curated specifically for training sparse neural networks. It proposes annealed \textit{activation smoothing} and decaying \textit{skip connection injection}, as well as a \textit{layer-wise rescaled initialization}. Code is provided for each of these methods. 

% \subsubsection{Other Training Techniques}
\textbf{RigL.} In the paper "Rigging the Lottery: Making All Tickets Winners" (\textit{RigL}) \cite{evci2021rigging}, Evci et al. design a training procedure that dynamically updates the sparse network topology, resulting in a fully-trained, sparse model in a single training stage. 
This methodology has been shown to accomplish \textbf{state of the art} on sparse training, making it a formidible baseline to compare against the pipeline that we develop. RigL operates by growing significant connections and pruning insignificant ones many times throughout the course of training.
\paragraph{}
\section{Methodology}
In this paper, we build an end-to-end pipeline for training sparse neural networks that is based on the training-based pruning paradigm. Given a sparsity parameter $s$ denoting the desired sparsity level ($s$ is the proportion of nonzero parameters to total parameters), we proceed as follows:
\begin{enumerate}
    \item In the \textbf{pre-pruning} stage, we train a model $f(x, \theta)$ to convergence using the methods detailed in \ref{section: prepruning}.
    \item Using the trained model $f(x, \theta)$, we apply \textit{magnitude-based pruning} on the model parameters $\theta$ in order to construct a mask $M \in \{0, 1\}^{|\theta|}$ such that $||M||_0 \leq s |\theta|$. 
    \item In the \textbf{post-pruning} stage, given the sparse mask $M$, we train the sparse model $f(x, \theta \odot M)$ to convergence using the methods detailed in \ref{section: postpruning}, resulting in a fully-trained, sparse model.
\end{enumerate}
This pipeline has the goal of pruning and network sparsity in mind from the beginning to the end, with each design decision curated to this specific setting. Furthermore, most prior approaches tend to seek undamaged performance for sparsity settings on the order of $s \sim 0.1$. We aim to push that an order of magnitude lower.

\subsection{Pre-Pruning Training}
\label{section: prepruning}
The end goal of the pre-pruning training stage is a neural network that is conducive to magnitude-based pruning. As such, we wish to deviate from the traditional deep learning training pipeline in ways that will help enforce sparsity of the model, measured by weight magnitude. Two such techniques will be useful for this purpose.
\subsubsection*{ZerO Initialization}
Normally, neural networks are initialized randomly with Gaussian weights sampled from $\mathcal{N}\left(0, \frac{C}{n_\ell}\right)$ for some constant $C$ with $n_\ell$ as either the fan-in or fan-out number. Clearly, this initialization begins with all weights having the same magnitude in expectation. Also, using the setting of a multilayer perceptron as a schematic example, each weight matrix initialized in this way is full-rank. As such, this initialization scheme (which is designed for general training) is not immediately specialized for the goal of training a network to be sparse. It then makes sense to apply an initialization scheme that \textit{is} specialized for such a task. For this purpose, we use the recently developed \textit{ZerO} initialization \cite{zhao2022zero}, which we detail below for linear layers. First, we define the types of matrices with which this initialization is built, following the notation in \cite{zhao2022zero}.
\begin{definition}[Partial identity matrix]
Let $\mathbf{I}_n$ denote the $n \times n$ identity matrix, and let $\mathbf{0_{r, c}}$ denote the $r \times c$ zero matrix. 
For given dimensions $r, c \in \mathbb{N}$, the $r \times c$ partial identity matrix $\mathbf{I}^*_{r, c}$ is defined as$$\mathbf{I}^* := \begin{cases}
    (\mathbf{I}_r, \mathbf{0}_{r, c - r}) & r < c \\
    \mathbf{I}_r & r = c \\
    (\mathbf{I}_c, 0_{c, r-c})^T & r > c
\end{cases} \in \mathbb{R}^{r \times c}$$
  
\end{definition}
\begin{definition}[Hadamard matrix]
    Define $\mathbf{H}_0 := [1]\in \mathbb{R}^{1 \times 1}$ to be the $0^{th}$ Hadamard matrix. Then, for all $m \in \mathbb{N}$, we define the $n^{th}$ Hadamard matrix recursively by
    $$\mathbf{H}_m := \begin{bmatrix}
        \mathbf{H}_{m-1} & \mathbf{H}_{m-1} \\ \mathbf{H}_{m-1} & -\mathbf{H}_{m-1}
    \end{bmatrix} \in \mathbb{R}^{2^m \times 2^m}$$
\end{definition}
Now, the \textit{ZerO} initialization algorithm for a multilayer perceptron is as follows (the algorithm is very similar for a convolutional layer, except that we treat the kernel as a $2 \times 2$ matrix by only initializing the diagonal):
\begin{algorithm}[H]
\caption{\textit{ZerO} initialization for MLPs.}\label{alg:cap}
\begin{algorithmic}
\Require A MLP with depth $L$ and layer widths $n_0, ..., n_L$, where $n_0$ is the input dimension and $n_L$ is the output dimension. Let $W^{(l)} \in \mathbb{R}^{n_l \times n_{l+1}}$ denote the $l^{th}$ weight matrix for $0 \leq l < L$.
\For{$l \in \{0, ..., L - 1\}$}
\If{$n_{l} = n_{l + 1}$}
    \State $W^{(l)} \gets I_{n_l}$
\ElsIf{$n_{l} < n_{l + 1}$}
    \State $W^{(l)} \gets I_{n_l, n_{l+1}}^*$
\ElsIf{$n_{l} > n_{l + 1}$}
    \State $W^{(l)} \gets cI_{n_l, 2^m}^* H_m I_{2^m, n_{l+1}}^*$, where $m := \lceil \log_2(n_l)\rceil$ and $c := 2^{-(m-1)/2}$
\EndIf
\EndFor
\end{algorithmic}
\end{algorithm}
It can be shown (see Lemma 1 in \cite{zhao2022zero}) that the weight matrices $W^{(l)}$ that we attain has a bounded rank on initialization. Furthermore, it is empirically shown that the weight matrices that are learned from this initialization monotonically increase in rank to their final state. This means that this initialization is well suited to learning low-rank transformations, which makes it conducive to sparisification via magnitude-based pruning (we expect a low rank matrix to have a few large elements along several directions).

\subsubsection*{Learning a Sparsity Mask}
In addition to the initialization presented above, we will also learn a sparsity mask throughout the pre-pruning stage. Precisely, we keep track of parameters $\theta$ of the network and a mask $M \in \{0, 1\}^{|\theta|}$. We interpret each element $M_i$ of $M$ to be a Bernoulli random variable determining whether the parameter $\theta_i$ will be used during each forward pass during training; in other words, $1 - M_i$ is the probability that $\theta_i$ is zeroed out during the forward pass. This will assign a sparse structure to the model immediately, and (if we allow gradients to backpropagate through the masking procedure) training of the weights $\theta$ will proceed as normally. However, backpropagation through this network will also train the mask $M$ in such a way that minimizes the loss. If we apply proper regularization, we can promote masks $M$ that have the two desired properties of sparse masks.
\newline \newline
More thoroughly, let $\ell(\widehat{y}, y)$ be the differentiable loss function that measures performance. Initialize $M$ to be $1/2$ in all entries. Define the regularized loss at a data point $(x, y)$ to be
$$L_{reg}(x, y) := \ell(f(x, \theta \odot M), y) + \lambda_1 \sum_{i=1}^{|\theta|} M_i(1 - M_i) + \lambda_2\sum_{i=1}^{|\theta|} M_i,$$
where $f(x, \theta)$ is the output of the network for input $x$ with parameters $\theta$, $M$ is the mask, and $\lambda_1, \lambda_2$ are regularization hyperparameters. Note that the $\lambda_1$ term is minimized when $M$ is composed of only 0's and 1's; this causes the optimization to prefer bimodal masks that are confident. The $\lambda_2$ term forces the optimization to prefer masks with more 0's and less 1's. Together, the added regularization terms enforce property 1 of sparse masks: that they are sparse. Meanwhile, the $\ell(f(x, \theta \odot M), y)$ term requires that the masked model's inference performs well.
\newline \newline
This scheme to learn a sparsity mask will ensure that the mask befits the model's weights, as they were jointly optimized. After pre-pruning training, we set $\theta \gets \theta \odot M$, such that magnitude-based pruning preferentially selects along the mask $M$ that we just learned.
\subsection{Post-Pruning Training}
After pruning, we now have a sparse model $f(x, \theta \odot M)$ that we wish to train to convergence (by this, we mean that only the elements $\{\theta_i : M_i \neq 0\}$ get trained). 
We apply several optimizations to the post-pruning training scheme to improve the training of this model. As one of the main difficulties of training sparse models lies in handling discontinuity and sharp transitions, most of these techniques encourage initial smoothness of the optimization trajectory.
\subsubsection*{Label Smoothing}
Label smoothing is a well-known technique that modifies the target labels by interpolating them with the uniform label. More specifically, let $K$ be the number of classes and let $\alpha$ be the label smoothing parameter. Then, for a given target label $y$, we produce the modified target label
$$\widetilde{y} := (1 - \alpha) \cdot y + \alpha \cdot \frac{1}{K}$$
This procedure smooths the labels (hence the name) against the uniform distribution. When cross-entropy is applied on top of this technique, the training optimization prefers smoother logit prediction as opposed to overconfident guesses. 
\newline \newline
Motivated by this technique, and following prior works, we apply a decayed label smoothing to the post-pruning training stage. By this, I mean that given hyperparameters $\alpha_{0}$ and a decay time $T_\alpha$ (measured in epochs), we set $\alpha$ for each epoch $t \in \{0, ..., T_\alpha\}$ to be
$$\alpha_t := \alpha_0 \left(1 - \frac{t}{T_\alpha}\right)$$
As such, $\alpha$ decays linearly from $\alpha_0$ to 0 in $T_\alpha$ epochs. This ensures that at the beginning of the post-pruning stage, the optimization trajectory begins smoother but eventually behaves normally (i.e. after $T_\alpha$ epochs, the loss is computed without any label smoothing). 
\subsubsection*{Soft Activations}
Following the work of \cite{jaiswal2022training} and others, we make use of a smoothed version of the ReLU activation in place of ReLU. This is because the gradient discontinuity of ReLU at 0 makes the optimization trajectories of sparse models sharp and tricky. To combat this, we apply a similar trick as we did in label smoothing. In particular, we define the parameterized Swish activation $\verb|PSwish|$ to be given by
$$\verb|PSwish|(t) := t \cdot \verb|sigmoid|(\beta t) = \frac{t}{1 + e^{-\beta t}}$$
for a temperature parameter $\beta$. Note that when $\beta = 0$ we have the (rescaled) identity, but as $\beta \rightarrow \infty$ we approach ReLU. This "soft activation" has continuous gradients and induces a smoother optimization trajectory, and so we would like to initially soften the activation and then increase $\beta$ during training. To this end, we define an initial $\beta_0$ and decay time $T_\beta$ (measured in epochs) as hyperparameters, and for each epoch $t \in \{0, ..., T_\beta\}$ we set
$$\beta_t := \beta_0 \left(1 - \frac{t}{T_\beta}\right) + \beta_{max} \frac{t}{T_\beta}$$
\subsubsection*{Soft Skip Connections}
Another point of failure of sparse network training in the initial phase is that, because of the sparse topology of the network, information propagation through the forward and backward passes can be halted or forced to pass through weird paths. Since we have removed a majority of the possible paths for gradients to flow, we would like to create some artificial ones in the beginning of the post-pruning training stage. To accomplish this, we create skip connections, as is done in \cite{jaiswal2022training}. Between each constituent block of the network, we add a weighted residual connection, parameterized by $\gamma$. More precisely, if $z$ is the intermediate tensor being passed into a block $g(\cdot; \theta_g)$, then we modify the output of the block to be
$$\widetilde{g}(z, \theta_g) := g(z, \theta_g) + \gamma \cdot z$$
(If the dimensions of $g(z, \theta_g)$ and $z$ do not add up, we simply resize and apply interpolation). As always, we introduce hyperparameters $\gamma_0, T_\gamma$ and for each epoch $t \in \{0, ..., T\}$ we define
$$\gamma_t := \gamma_0\left(1 - \frac{t}{T_\gamma}\right)$$
and use $\gamma_t$ as the parameter for that epoch. This allows the network to initially optimize as though it had many skip connections and allow gradients to flow reasonably. However, later on in training (after $T_\gamma$ epochs), we recover the intital topology of the model so that we don't sacrifice final performance.
\label{section: postpruning}

% In this paper, we choose to investigate magnitude-based pruning so that our results are comparable with those seen in \cite{lth} and related works. Training-free methods will be used as baselines to compare our methods against. It is important, however, to keep in mind that our methods will make use of pre-pruning training and training-based magnitude pruning, while the above methods only make use of initialization information. 

% \newline \newline
% In this paper, we design various well-motivated combinations of the above complementary techniques and assess their performance against each other and our baselines.
\section{Experiments}
\subsection{Settings}
\textbf{Baselines.} We compare the proposed methodology against training-free pruning pipelines (i.e. random, SNIP, GraSP, and SynFlow \cite{snip, grasp, synflow}), as well as against RigL, a state of the art sparse training pipeline that does not use a pruning phase. For RigL, we use the suggested hyperparameters. 
\newline \newline
\textbf{Datasets.} We evaluate our proposed pipeline in two multi-class classification settings: 
\begin{enumerate}
    \item on the MNIST dataset with a fully-connected 6-layer MLP with 100 hidden units per layer
    \item on the CIFAR-10 dataset with a VGG-16 model
\end{enumerate}
\textbf{Hyperparameters.} We set the following training hyperparameters for all experiments:
\begin{multicols}{2}
\begin{itemize}
    \item \verb|batch_size| = 256
    \item \verb|learning_rate| = 0.001
    \item \verb|weight_decay| = $1 \times 10^{-6}$
    \item $\verb|num_epochs| = \begin{cases}
        20 & \text{MNIST} \\
        100 & \text{CIFAR-10}
    \end{cases}$
    \item We use the Adam optimizer.
\end{itemize}
\end{multicols}
Furthermore, for the hyperparameters that are introducted in our methodology, for the MNIST experiments we set:
\begin{multicols}{2}
    \begin{itemize}
        \item $\lambda_1 = 0.001$
        \item $\alpha_0 = N/A$
        \item $\beta_0 = N/A$
        \item $\gamma_0 = N/A$
        \item $\lambda_2 = 0.05$
        \item $T_\alpha = N/A$
        \item $T_\beta = N/A$
            \item $T_\gamma = N/A$
            \end{itemize}
    \end{multicols}
    and for the CIFAR-10 experiments we set:
\begin{multicols}{2}
    \begin{itemize}
        \item $\lambda_1 = 0.01$
        \item $\alpha_0 = 0.3$
        \item $\beta_0 = 1.0$
        \item $\gamma_0 = 0.4$
        \item $\lambda_2 = 0.1$
        \item $T_\alpha = 50$
        \item $T_\beta = 60$
            \item $T_\gamma = 60$
            \end{itemize}
    \end{multicols}
    The above hyperparameters were determined on a validation dataset. (By $N/A$ for the MNIST experiments, we mean that we saw no observable differences when turning on label smoothing, soft activations, nor soft skip connections, and so we leave them off for the experiments below). 
\subsection{Results}
In the plots that follow, the following information is relevant:
\begin{itemize}
\item $s$ is the sparsity parameter that dictates the proportion of parameters that is left nonzero after sparsification. The related "compression" value, which is sometimes used to parameterize this, is related by $compression = -\log_{10}(s)$.
    \item $\verb|training-free|$ denotes the result of the pruning pipeline where the best training-free pruning technique was used, and then the resulting sparse model was trained for \verb|num_epochs| epochs; in other words, this plots the corresponding values for $\max \{\verb|random|, \verb|SNIP|, \verb|GraSP|, \verb|SynFlow|\}$ for each experiment (it was usually SNIP).
    \item \verb|RigL| denotes the result of training for $\verb|num_epochs|$ using the \textit{RigL} training scheme with the desired sparsity $s$.
    \item All other runs followed the procedure of 
    $$\text{train for num\_epochs $\rightarrow$ magnitude-based pruning $\rightarrow$ train for num\_epochs}$$
    with the combination of techniques given in the legend.
    \item We use \verb|ToST| to abbreviate all of the post-pruning training optimizations, since we only tried turning them on together.
    \item Dashed lines are used to indicate baselines.
    \item \verb|vanilla| is used to indicate vanilla magnitude pruning.
\end{itemize}
\begin{figure}[H]
    \centering
\includegraphics[width=5in]{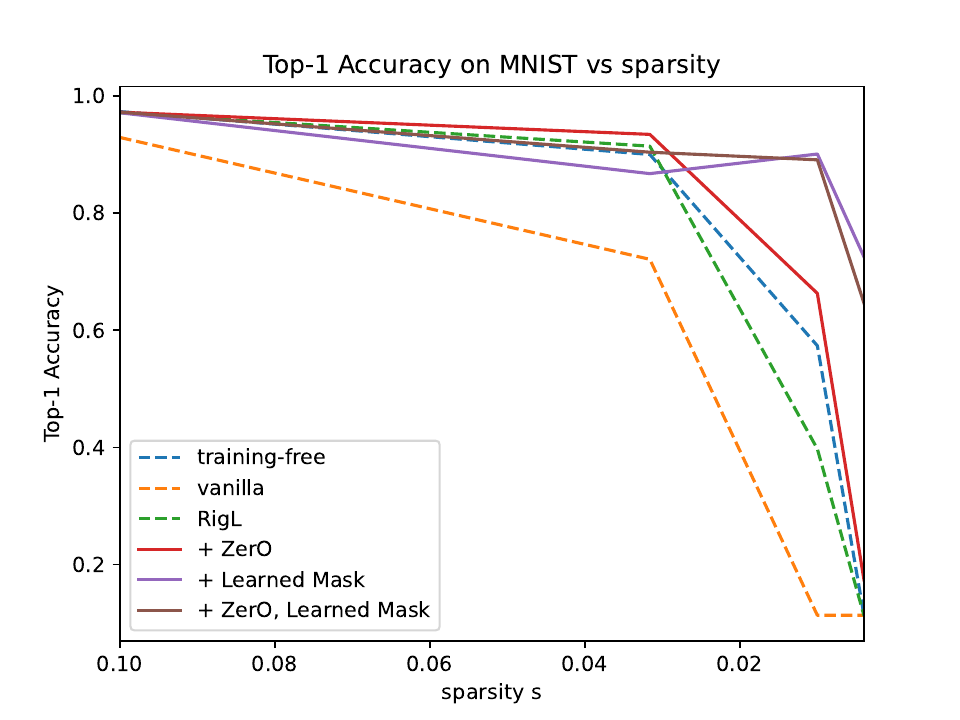}    \caption{Top-1 accuracies on the MNIST dataset using the MLP architecture. We see that our proposed methods offer improvement over the baselines, with the learned mask having the largest impact. Furthermore, the addition of the post-pruning training optimizations seems to help when applied, but very barely.}
    \label{fig:mnist}
\end{figure}
\begin{figure}[H]
    \centering
\includegraphics[width=5in]{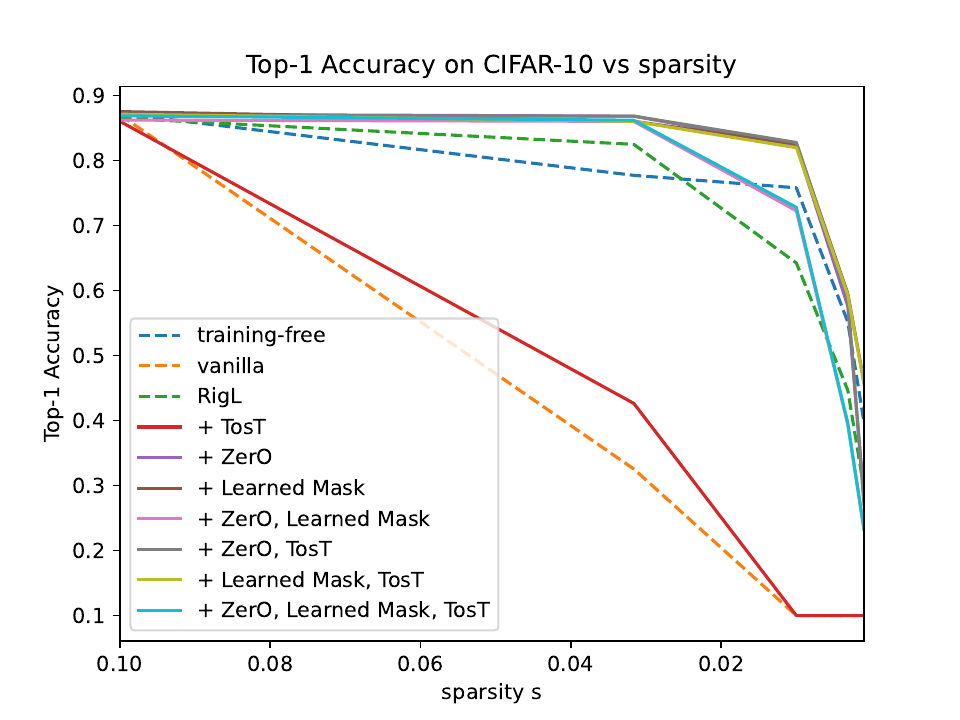}    \caption{Top-1 accuracies on the CIFAR-10 dataset using the VGG-16 architecture. We see that our proposed methods offer improvement over the baselines, with the learned mask having the largest impact. Furthermore, the addition of the post-pruning training optimizations seems to help when applied, but very barely.}
    \label{fig:cifar}
\end{figure}

\begin{table*}[ht]
\center
\begin{tabular}{lccccc}
 & \multicolumn{1}{c}{$s = 10 \%$} & \multicolumn{1}{c}{$s = 3.2 \%$} &\multicolumn{1}{c}{$s = 1 \%$} &\multicolumn{1}{c}{$s = 0.32 \%$}\\
\midrule
\verb|training-free| &\textbf{97.3} & 93.2 & 57.0 &11.4 \\
\verb|vanilla| &88.9 & 75.2 & 11.4 &11.4 \\
\verb|RigL|  &97.1 & 92.8 & 40.1 &11.4 \\
\midrule
\verb|+ ZerO|  &97.2 & \textbf{97.1} & 66.2 &17.9 \\
\verb|+ Mask| &97.0 & 96.5 & 88.9 &\textbf{73.5} \\
\verb|+ ZerO, Mask|  &97.2 & 87.8 & \textbf{89.0} &65.0 \\
\bottomrule
\end{tabular}\center
\label{table:mnist}
\caption{Reported Top-1 Accuracies of MLP architecture trained on the MNIST dataset. We find that the \textit{ZerO} initialization and learned masks seem to have the largest effects on pruned accuracy for sparsities on the order of 3\%, but for anything less it is the learned mask that has the largest impact.}
\end{table*}

\begin{table*}[ht]
\center
\begin{tabular}{lccccc}
 & \multicolumn{1}{c}{$s = 10 \%$} & \multicolumn{1}{c}{$s = 3.2 \%$} &\multicolumn{1}{c}{$s = 1 \%$} &\multicolumn{1}{c}{$s = 0.32 \%$} & \multicolumn{1}{c}{$s = 0.1 \%$}\\
\midrule
\verb|training-free| &87.3   & 77.7  &  75.8 & 55.2 &39.7  \\
\verb|vanilla| &87.1   & 34.0  &  10.0 & 10.0 &10.0  \\
\verb|RigL| &86.5   & 82.4  &  64.2 & 44.7 &29.5  \\
\midrule
\verb|+ ZerO| &87.1   & 86.7  &  82.1 & 57.4 &27.6  \\
\verb|+ Mask| &\textbf{87.5}   & 86.0  &  81.1 & 59.1 &\textbf{45.7}  \\
\verb|+ ToST| &87.2   & 39.1  &  10.0 & 10.0 &10.0  \\
\verb|+ ZerO, ToST| &86.9   & \textbf{86.8}  &  \textbf{83.1} & 59.4 &27.0  \\
\verb|+ Mask, ToST| &87.1   & 86.0  &  \textbf{83.1} & \textbf{60.1} &42.2  \\
\verb|+ All| &86.8   & 86.2  &  72.4 & 39.6 &22.8   \\
\bottomrule
\end{tabular}\center
\label{table:cifar}
\caption{Reported Top-1 Accuracies of VGG-16 architecture trained on the CIFAR-10 dataset. We find that the \textit{ZerO} initialization and learned masks seem to have the largest effects on pruned accuracy.}
\end{table*}
The above results demonstrate several things. Firstly, our combined approach does significantly better than the baselines when the models become very sparse. Indeed, we are able to retain full performance with only 3\% of the parameters unpruned, which is better than both the training-free pruning baselines and \textit{RigL}. At $\leq 1\%$ of the parameters left, our methodology seems to dominate. 
\newline \newline
There are, however, some interesting and worrisome trends to note. Firstly, it appears the \textit{ZerO} initialization and learned masks don't play well with each other; we discuss this in Section \ref{section: discussion}. Furthermore, we find that the gap between our pipeline and the baselines is much larger in MNIST than in CIFAR-10. Perhaps things look quite different if we continue to scale the experiments; this would be interesting to investigate. Lastly, the combination of label smoothing, soft activations, and soft skip connections seems to have a \textit{very small positive effect}, smaller even than what is reported in \cite{jaiswal2022training} (though they weren't testing at such small sparsity levels). 
% Sparsity & Dataset & \verb|training-free| & \verb|vanilla| & \verb|RigL|  & \verb|+ZerO|& \verb|+Mask|& & \verb|+ToST|& \verb|+all|

\section{Discussion}
\label{section: discussion}

Our combined methodology appears to beat the well-established previous methods and also a fairly new state of the art in very sparse settings. Interestingly, it appears that much of the heavy lifting is done by the learnable masks, and so further research in this direction seems fitting. 
\newline \newline
As mentioned earlier, the results indicate that the \textit{ZerO} initialization is combined with the learnable mask do not combine in a way that improves final performance. Upon further examination, this is indeed the case for the following reason: on initialization and early in the pre-pruning training stage, the gradients of the cross-entropy with respect to the mask are wildly undirected. As such, the $\lambda_2$ regularization term that attempts to sparsify the mask sends the entire mask to 0 and training collapses irreparably; this happens for any $\lambda_2 > 0$. If we run the same experiment but instead set $\lambda_2 = 0$, training no longer collapses, but instead the learned mask is all 1's and we recover the same pruned performance that \textit{ZerO} accomplishes on its own. Further work in investigating the interplay between these two methods seems promising, as they both individually offer fantastic improvement and appear to operate in complementary ways. Perhaps the scheme with which we initialize the mask could be modified to fit these two techniques together?
\newline \newline
Lastly, the results for very small sparsities are tantalizing; for a sense of scale, recall that 1\% sparsity of VGG-16 is 147k parameters remaining out of the original 14.7M. All of the works cited in this paper do not attempt sparsities this small, and so it is useful knowledge just to understand how pruning looks at this scale. That being said, it would be very interesting to know more about how methods such as the proposed pipeline scale to larger models and more complex settings. Operating with such small sparsities isn't all that useful for VGG-16 (the difference between 1 and 3\% is not much), but such sparsities would be crucially important if they could be usefully reached in much larger models. In particular, it is important to see if the same techniques apply to large language models and large models that mix convolution and attention, since this can be where the pruning of parameters finds biggest impact. Furthermore, in the \textit{ZerO} initialization paper \cite{zhao2022zero} the author claims to have reached similar results for the initialization of attention layers, though I have not been able to replicate this nor find an implementation. If the results of our methodology here carry over to attention layers (which have inference time $n^2$), the inference footprint rewards of our increased training efforts could be very large. 

\section{Acknowledgements}
Thank you so much to Professor Kai Li and to Yushan for a fantastic course! I loved the opportunity to learn about a topic I didn't know all that much about, and the paper selection was incredible. Thank you :)

\bibliographystyle{unsrt}
\bibliography{ref}
\end{document}